\documentclass[letterpaper]{article} 
\usepackage{aaai25}  
\usepackage{times}  
\usepackage{helvet}  
\usepackage{courier}  
\usepackage[hyphens]{url}  
\usepackage{graphicx} 
\usepackage{natbib}  
\usepackage{caption} 
\usepackage{amsmath}
\usepackage{algorithm}
\usepackage{algorithmic}

\usepackage{newfloat}
\usepackage{listings}
\DeclareCaptionStyle{ruled}{labelfont=normalfont,labelsep=colon,strut=off} 
\floatstyle{ruled}
\newfloat{listing}{tb}{lst}{}
\floatname{listing}{Listing}
\title{Structured Reasoning for Fairness: A Multi-Agent Approach to Bias Detection in Textual Data}

\author {
    Tianyi Huang\textsuperscript{\rm 1, \rm 2}\thanks{Primary Author and Corresponding Author},
    Elsa Fan\textsuperscript{\rm 2}
}

\affiliations {
    \textsuperscript{\rm 1}App Inventor Foundation\\
    \textsuperscript{\rm 2}App-In Club\\
    tianyi@appinventorfoundation.org, elsa@appinclub.org
}

\begin{document}

\maketitle

\begin{abstract}
From disinformation spread by AI chatbots to AI recommendations that inadvertently reinforce stereotypes, textual bias poses a significant challenge to the trustworthiness of large language models (LLMs). In this paper, we propose a multi-agent framework that systematically identifies biases by disentangling each statement as fact or opinion, assigning a bias intensity score, and providing concise, factual justifications. Evaluated on 1,500 samples from the WikiNPOV dataset, the framework achieves 84.9\% accuracy—an improvement of 13.0\% over the zero-shot baseline—demonstrating the efficacy of explicitly modeling fact versus opinion prior to quantifying bias intensity. By combining enhanced detection accuracy with interpretable explanations, this approach sets a foundation for promoting fairness and accountability in modern language models.

\end{abstract}

\section{Introduction}

Words hold immense power in shaping perceptions, influencing social exchanges, and driving decision-making processes. In the era of large language models (LLMs), this power is amplified, as automated systems now participate in generating and interpreting large volumes of textual data at unprecedented scales \cite{devlin2019, vaswani2023}. The reach of LLMs extends from assisting medical diagnoses and legal contract analysis to moderating online content and supporting educational tools \cite{10.1371/journal.pdig.0000662}. Despite their remarkable influence and capabilities, these systems often inherit the biases embedded in their training data, risking the perpetuation of harmful stereotypes or discriminatory language \cite{gallegos2024, Mei2023}. Equally problematic, subtle subjectivity and skewed phrasing may pass unnoticed, exposing end-users to outputs that inadvertently frame narratives in ways misaligned with fairness \cite{10.1145/3442188.3445922}. These challenges emphasize the urgent need for bias detection methods that not only identify problematic content but also clarify how and why biases arise \cite{li2024}.

\begin{figure}[ht]
    \centering
    \includegraphics[width=0.35\textwidth]{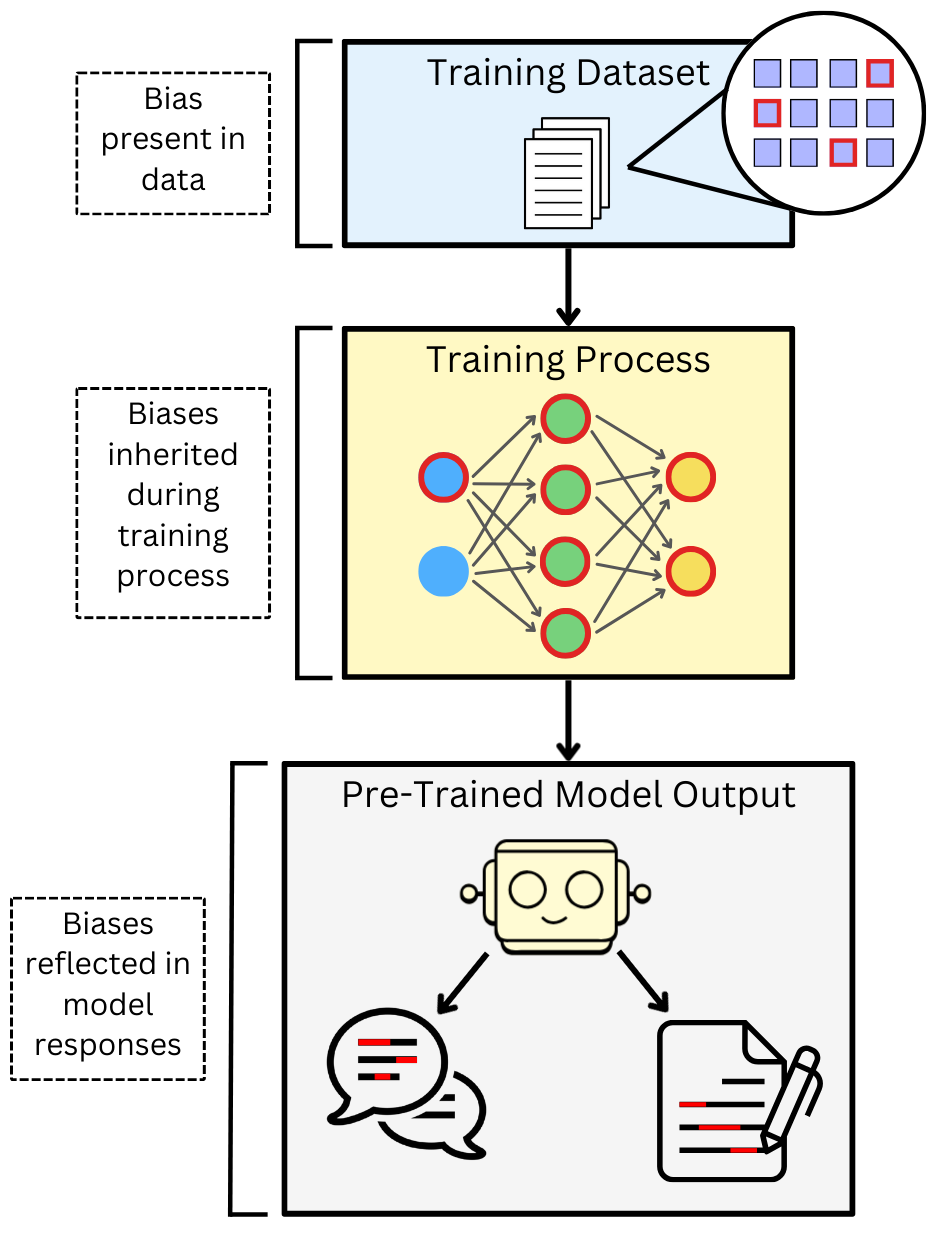}
    \caption{An illustration of how biases present in a training dataset can be inherited by an AI model during training and reflected in the model's responses, potentially compromising objectivity.}
    \label{fig:bias_detection_intro}
\end{figure}

Existing approaches to bias detection and mitigation often rely on static lexicons or predefined rules, which fail to capture the nuances of emerging or context-dependent biases \cite{husse2022, webster2021}. Another limitation is that these methods would lack explainability, reducing transparency in AI-driven decisions \cite{petkovic2022}. Moreover, certain methods simply mask biased terms or phrases without providing insights into the broader social or factual underpinnings of the bias \cite{dev2019}. Consequently, there remains a gap in the literature for frameworks that integrate factual verification, subjective analysis, and transparent explanations. 

In this paper, we introduce a multi-agent framework that aims to tackle these shortfalls through systematically detecting bias and opinionated language in textual data through a structured reasoning process. Our pipeline includes:
\begin{enumerate}
    \item A checker agent that classifies a statement as factual or opinion-based, removing ambiguity in subsequent analyses.
    \item A validation agent that measures the intensity of bias in opinionated statements using a validity scoring mechanism, ensuring both subtle and overt biases are captured.
    \item A justification module that provides explanations of the final classification, promoting better interpretability and transparency.
\end{enumerate}

Beyond improving bias detection, our work contributes to broader efforts in creating accountable and socially responsible AI: by integrating this holistic approach, it holds promise for real-world deployments where unbiased AI outcomes are imperative—ultimately advancing the goal of developing AI technologies that uplift rather than undermine societal well-being.

\section{Related Works}

Research on bias in Natural Language Processing (NLP) has evolved considerably over the last decade, driven by growing concerns over computational models often reflecting and amplifying existing societal prejudices. Early approaches for addressing this issue largely focused on debiasing static word embeddings, as demonstrated by Bolukbasi et al. (NIPS 2016), who identified systematic gender biases in vector representations and proposed geometric alignment techniques to mitigate them \cite{bolukbasi2016}. While these initial efforts effectively highlighted the pervasiveness of stereotyping in word embeddings, they addressed only limited linguistic contexts and were insufficient in capturing the subtleties of contextualized language models.

Subsequent work expanded the focus to contextual embeddings. Zhao et al. illustrated how transformer-based models inadvertently perpetuate gender and racial biases across various NLP tasks, emphasizing the potential adverse consequences for downstream applications \cite{zhao2019}. Efforts to measure and quantify bias in contextual representations often rely on carefully designed benchmarks and diagnostic tests, such as the StereoSet and CrowS-Pairs datasets, which reveal performance disparities correlated with sensitive attributes \cite{nadeem2020, nangia-etal-2020-crows}. However, purely quantitative evaluation methods can frequently overlook more nuanced forms of bias—particularly statements that embed subtle value judgments rather than including explicit biases \cite{zhao2025}.

A second line of inquiry examines explainability and interpretability as prerequisites for credible bias detection. Ribeiro et al. (KDD 2016) proposed Local Interpretable Model-Agnostic Explanations (LIME) to help end-users understand and trust classifier decisions \cite{ribeiro2016}. Lundberg and Lee (NIPS 2017) introduced SHAP (SHapley Additive exPlanations), a unified framework for interpreting predictions across a variety of models \cite{lundberg2017}. While these techniques demystify model outputs by highlighting salient tokens or phrases, they do not always pinpoint the origin of biases in training data or account for the degree to which an entire statement might be skewed.

More recent research ventures into multi-faceted bias detection that integrates social context, factuality checks, and user feedback loops. For instance, Field and Tsvetkov (2020) explored unsupervised methods for classifying gender in text, emphasizing the importance of considering contextual factors in bias detection \cite{field2020}. In parallel, integrated pipelines for bias analysis—e.g., evaluating explicit sentiment, measuring harmful stereotypes, and quantifying subjectivity—have proven beneficial in tasks such as moderated content filtering and hate speech detection \cite{garg2022, liu2024, aoyagui2024}. Still, many existing tools provide fragmented insights; some focus solely on word-level biases, while others rely on rigid rules that fail to adapt to evolving language trends. Moreover, transparency often remains insufficient: users may see a biased word flagged but lack an explanation grounded in factual evidence or logical reasoning.

Although bias detection in NLP has progressed through the advent of various approaches, current solutions fail to address three main problems: quantitative methods struggle in identifying nuanced biases, interpretability methods face difficulty in determining the full extent of bias in statements, and approaches involving integrated evaluation focus solely on subjective components. Consequently, these methods become ineffective in detecting statements that appear factual yet contain subtle biased language, as a deeper analysis rooted in factual evaluation and contextual understanding is required. Furthermore, these systems often fall short of producing valid and logical explanations for bias detection, hindering progress in ensuring full transparency for LLMs. To address these concerns, we propose a multi-agent framework that detects biases using a systematic approach. Unlike computational methods, our system determines implicit biases by distinguishing factual content from opinion-based text and quantifies varying degrees of bias to handle limitations present in the LIME and SHAP frameworks. Additionally, we also reduce the risks involved in analyzing subjectivity by focusing on classifying the nature of the statements themselves and evaluating their ability to be verified with evidence. These advances and the inclusion of justification responses place our framework as a solution for reinforcing fairness in AI systems. 

\section{Methodology}

\begin{figure*}[t]
\centering
\includegraphics[width=\textwidth]{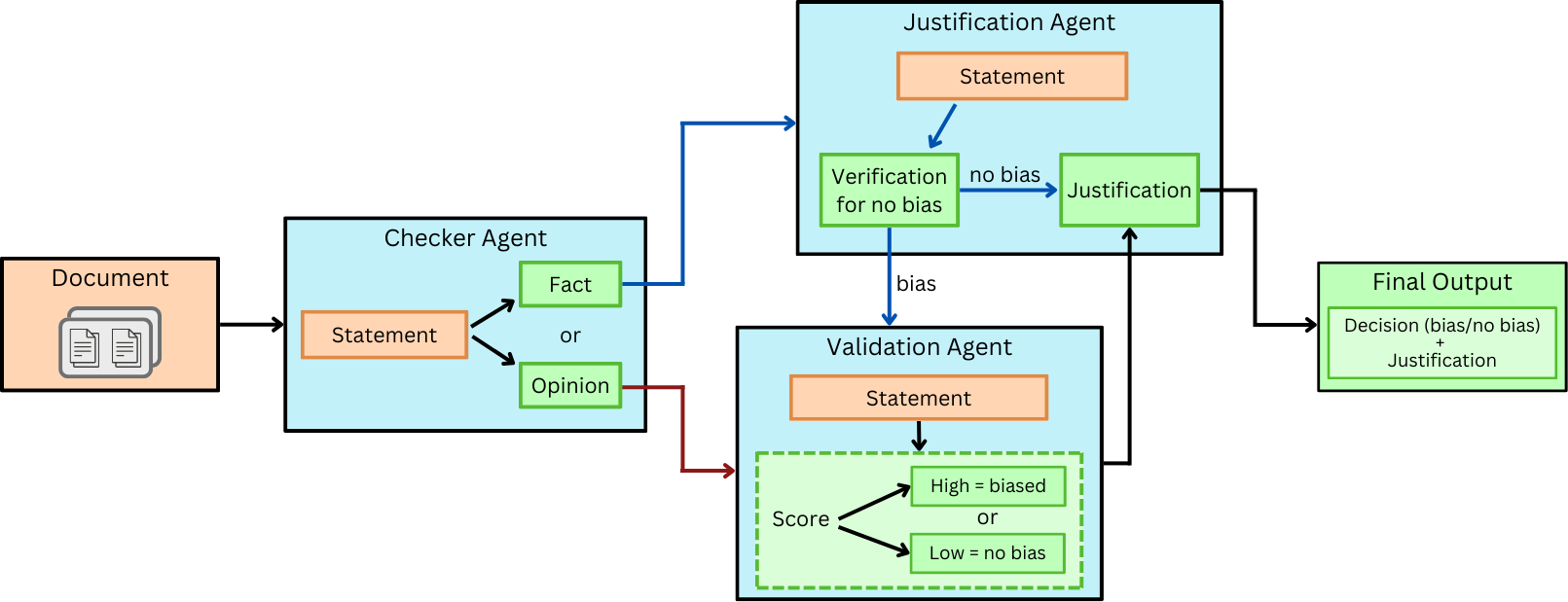}
\caption{Overview of the multi-agent bias detection pipeline. Text statements first enter a checker agent to be classified as \emph{fact} or \emph{opinion}. Factual statements are then verified by a justification agent for bias, while opinionated statements undergo evaluation by a validation agent. Finally, the system outputs a final decision (biased or unbiased) alongside a concise justification.}
\label{fig:bias_detection_pipeline}
\end{figure*}

\subsection{Checker Agent: Fact vs.\ Opinion Classification}
The initial step of our system is determining whether a statement is purely factual or contains subjective elements. We let \(S\) denote the statement and define a decision function:
\[
\mathrm{Decision}(S) =
\begin{cases} 
\text{FACT}, & \text{if $S$ is purely verifiable}, \\
\text{OPINION}, & \text{otherwise}.
\end{cases}
\]
A statement labeled as \emph{FACT} is expected to be completely objective and testable against empirical evidence, while any presence of interpretive or persuasive language triggers an \emph{OPINION} label. This initial filter ensures that the subsequent steps can be fitted to the specific nature of the statement.

\subsection{Handling Factual Statements}
If the checker agent outputs \emph{FACT}, the pipeline applies a minimal bias verification step to confirm that the statement’s language or presentation does not subtly introduce skew or partial framing. Specifically:
\begin{itemize}
    \item \textbf{Factual-Bias Verification:} Factual statements may occasionally exhibit bias through wording, emphasis, or selective omission. If the pipeline detects no bias here, it forwards the statement to the justification writing with a ``No Bias'' outcome.
    \item \textbf{Potential Bias Escalation:} If this initial check suggests that the factual statement may contain bias, it is routed to the validation agent for a more in-depth inspection. This approach conserves computational costs by avoiding unnecessary full-scale analysis in obviously unbiased factual cases.
\end{itemize}

\subsection{Validation Agent: Bias Scoring}
All statements labeled \emph{OPINION} by the checker agent, along with any factual statements flagged as potentially biased, are sent to the validation agent for full-scale analysis. Formally, we write:
\begin{equation}
    \mathrm{Validate}_\mathrm{bias}(S) = f_{\mathrm{LLM}}(S),
\end{equation}
where \( f_{\mathrm{LLM}} \) is a large language model tasked with assessing the extent of bias. For opinion statements, the agent looks for subjective or emotional language, strong value judgments, and imbalance in perspective. For escalated factual statements, it focuses on how factual content may be presented in a biased manner (e.g., emotive tone, selective emphasis). The validation agent then assigns a \emph{Bias Level} of \texttt{HIGH} or \texttt{LOW}. We interpret \texttt{HIGH} as a binary \(\text{predicted\_bias} = \text{True}\) and \texttt{LOW} as \(\text{predicted\_bias} = \text{False}\).

\subsection{Justification Agent}
Regardless of the validation outcome, a justification agent is called to generate a concise explanation of the verdict. This agent references the reasoning steps that led to either \emph{No Bias} or \emph{Bias}. Specifically,
\begin{itemize}
    \item \textbf{No-Bias Cases:} The justification emphasizes the statement’s objectivity and neutrality.
    \item \textbf{Biased Cases:} The justification pinpoints specific words, framing devices, or tones that caused the bias classification.
\end{itemize}
This interpretability step aims to allow for greater transparency, especially in high-stake applications where reliability is paramount.

\subsection{Final Output}
After passing through the above stages, the pipeline produces two pieces of information:
\begin{enumerate}
    \item \textbf{Binary Bias Classification:} ``Bias'' or ``No Bias.''
    \item \textbf{Justification:} A concise explanation detailing the reasons behind the classification.
\end{enumerate}
These outputs are stored in a .json file for subsequent metric calculations and potential use in applications that require auditability or further inspections.

\subsection{Implementation Details}
We implement our pipeline in a modular structure by making asynchronous calls to a large language model at each agent stage:
\begin{itemize}
    \item \textbf{Choice of LLM:} In our experiments, we primarily used \emph{GPT-4o} to perform classification, bias verification, and justification generation \cite{gpt4o}. Nonetheless, the design can be integrated with other LLMs as long as it follows the prompts and output formats.
    \item \textbf{Data Pipeline:} We randomly sample 1,500 labeled statements (biased and unbiased) from the WikiNPOV dataset, ensuring consistency via a fixed random seed \cite{Hube_2019}. Each statement travels asynchronously through the checker, bias-verification (if factual), validation, and justification steps, which support scalability in large datasets.
\end{itemize}

\section{Baseline Approach}
In addition to our multi-agent pipeline, we employ a \emph{zero-shot baseline} for comparative evaluation. This baseline directly prompts GPT-4o (or any other preferred LLM) to classify each statement as either "biased" or "unbiased" without employing specialized fact-opinion segmentation \cite{gpt4o}. The model operates with a single instruction focused solely on identifying bias in language or presentation, producing a one-word output per statement. This zero-shot approach serves as a valuable comparison for evaluating the pipeline's effectiveness in enhancing bias detection.

\subsection{Performance Metrics}
We measure the pipeline’s effectiveness by comparing predicted labels (\(\hat{y}\)) against the ground truth (\(y\)) on the sampled statements. Specifically, we compute the following:

We measure the effectiveness of both our multi-agent pipeline and the baseline by comparing their predicted labels \(\hat{y}\) to the ground truth \(y\) for each of the sampled statements. Specifically, we use standard metrics such as accuracy, precision, recall, and F1 score:

\begin{figure}[h]
    \centering
    \includegraphics[width=0.46\textwidth]{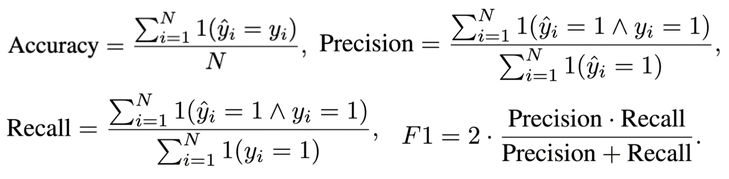}
\end{figure}

Here, \(N\) is the total number of evaluated statements, and $1(\cdot)$ is the indicator function. These metrics, complemented by detailed logs of final decisions (e.g., statement type, bias level, justification), enable full assessments of our methodology’s robustness.

\section{Results}

\subsection{Comparing Pipeline and Baseline}
We evaluate our \emph{Pipeline} (checker--validation--justification) against a \emph{Baseline} that relies on a single prompt to classify each statement as ``biased'' or ``unbiased.'' Both methods use GPT-4o as their underlying LLM. Table~\ref{tab:main-results} summarizes the results on the 1,500-statement sample from the WikiNPOV dataset, with performance metrics reported in percentages.

\begin{table}[h]
\centering
\begin{tabular}{lcccc}
\hline
\textbf{Method} & \textbf{Accuracy} & \textbf{Precision} & \textbf{Recall} & \textbf{F1 Score} \\ \hline
Baseline & 0.719 & 0.328 & 0.839 & 0.472 \\
Pipeline & 0.849 & 0.494 & 0.518 & 0.505 \\
\hline
\end{tabular}
\caption{Performance on a 1,500-statement subset of the WikiNPOV dataset (GPT-4o).}
\label{tab:main-results}
\end{table}

\paragraph{Statistical Significance.}
We conducted a two-proportion $z$-test to verify whether the accuracy difference between the two methods is statistically significant. Specifically, for the 1,500-statement test set, the Baseline correctly classifies approximately $1,079$ instances ($71.9\%$), whereas the Pipeline correctly classifies around $1,274$ ($84.9\%$). The resulting $z$-score is above $8.0$, yielding a $p$-value below $10^{-6}$. This indicates that our pipeline’s improvement in accuracy is both practically meaningful and statistically robust.

\paragraph{Confusion Matrices.}
Figures 3 and 4 represent the confusion matrices for the Baseline and Pipeline, respectively. The Baseline demonstrates a relatively high Recall (few \emph{false negatives}) but struggles with Precision, as evidenced by its numerous \emph{false positives}. In contrast, our Pipeline achieves a more balanced distribution of true positives and true negatives, thereby attaining a higher F1 Score and offering superior overall reliability.

\begin{figure}[h]
\centering
\includegraphics[width=0.95\columnwidth]{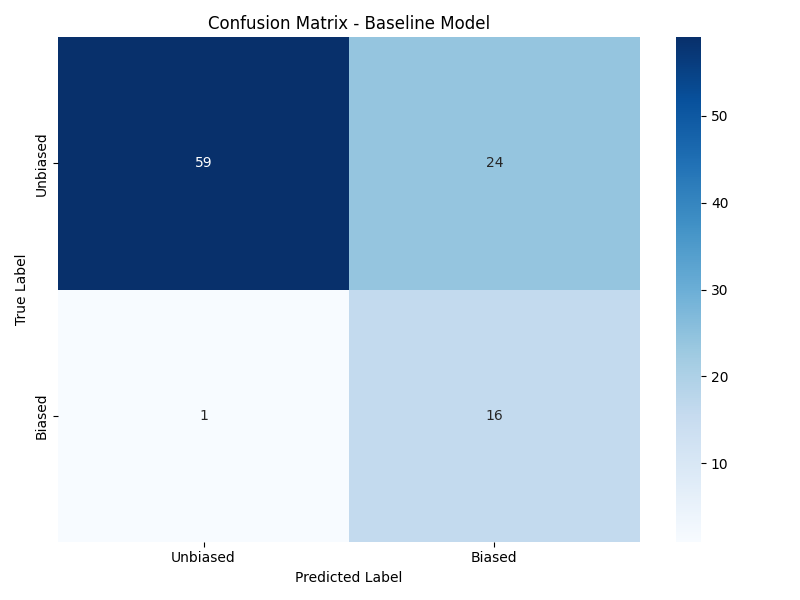}
\caption{Confusion Matrix for the Baseline on 100 WikiNPOV statements (GPT-4o).}
\end{figure}

\begin{figure}[h]
\centering
\includegraphics[width=0.95\columnwidth]{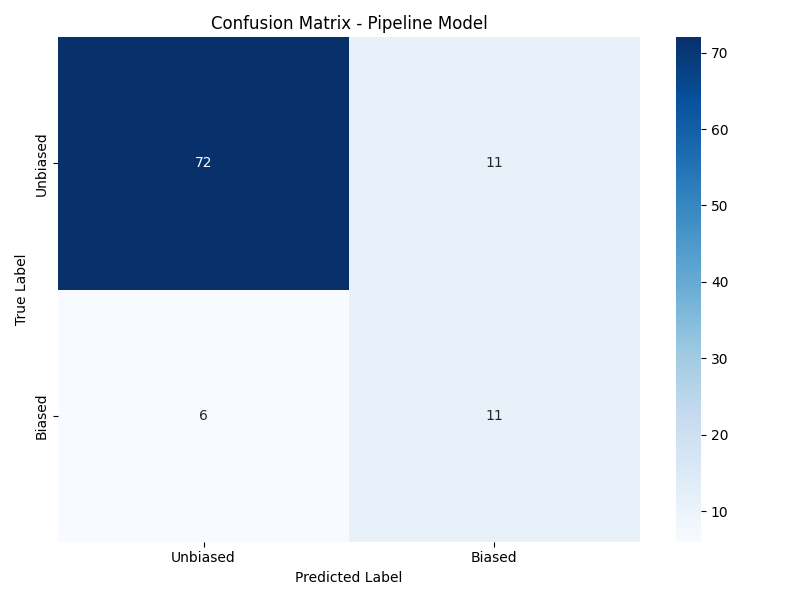}
\caption{Confusion Matrix for the Pipeline on 100 WikiNPOV statements (GPT-4o).}
\end{figure}

\subsection{Adaptability to Different LLMs}
To assess generality, we applied our Pipeline with three different state-of-the-art LLMs on a smaller 100-statement sample: GPT-4o, Claude 2.1, and Google Gemini 1.5 Flash \cite{gpt4o, claude2.1, gemini1.5}. Table 2 shows that although there is some variability (especially in Recall) when switching to Claude, the Pipeline remains operational and achieves around 80\% Accuracy in all settings. These findings show that our multi-agent design is not tightly coupled to one particular model and can be migrated to alternative LLMs.

\begin{table}[h]
\centering
\begin{tabular}{lcccc}
\hline
\textbf{LLM} & \textbf{Accuracy} & \textbf{Precision} & \textbf{Recall} & \textbf{F1 Score} \\ \hline
GPT-4o & 0.830 & 0.500 & 0.647 & 0.564 \\
Claude & 0.810 & 0.400 & 0.235 & 0.296 \\
Gemini & 0.780 & 0.696 & 0.780 & 0.736 \\
\hline
\end{tabular}
\caption{Performance of pipeline comparing three LLMs (100-statement subset), including GPT-4o, Claude 2.1, and Google Gemini 1.5 Flash.}
\label{tab:llm-comparison}
\end{table}

\subsection{Qualitative Insights}
While the Baseline model allows for only a single-word prediction for each statement, our Pipeline provides a structured JSON output containing fields such as \texttt{analysis}, \texttt{bias\_score}, and \texttt{justification}. As shown in the listing, the Baseline simply returns a label (\texttt{"biased"}) for the example statement. Notably, the Pipeline not only labels a statement as \emph{biased} or \emph{unbiased} but also explains why it made that decision. This interpretability is beneficial for domains where traceability and reliability are important.

\begin{listing}[h]
\caption{Comparing JSON Output of Baseline and Pipeline}
\begin{lstlisting}
  {
    "text": "A correct understanding...gravitation ",
    "true_label": 0,
    "predicted_label": 0,
    "raw_response": "biased"
  } # Baseline

  {
    "text": "A correct understanding...gravitation ",
    "true_label": 0,
    "predicted_label": 0,
    "statement_type": "opinion",
    "analysis": {
       "fact_check": null,
       "bias_score": high,
       "justification": "The statement shows bias by presenting general relativity as the only valid framework, which disregards alternative theories or interpretations. It lacks balance..."
     }} # Pipeline
\end{lstlisting}
\end{listing}

\subsection{Discussion}
The significant improvements in accuracy ($p < 10^{-6}$) validate the effectiveness of this multi-agent pipeline for bias detection tasks with a checker, validation, and justification agent. Additionally, the framework’s adaptable architecture allows the LLM model to be switched, making it extensible to new and emerging models.

\section{Limitations and Future Work}

\subsection{Limitations}
Although our framework shows promise in improving bias detection for textual data, several challenges should be considered:  
\begin{itemize}
    \item \textbf{Dataset Dependency:} The WikiNPOV dataset that our system is evaluated on consists of limited data that, while comprehensive, may not encompass all biases present in real-world situations \cite{Hube_2019}. The framework’s performance could vary when applied to domains or datasets with different patterns or contextual requirements, drawing the need for broader dataset evaluations.
    \item \textbf{Uncertainty in Classifying Facts and Opinions:} Our framework relies on a binary categorization of statements as either factual or opinion-based. Statements blending factual and opinionated elements may lead to misclassifications, reducing the accuracy of subsequent bias detection. This limitation underscores the need for a more nuanced classification mechanism capable of handling hybrid statements.
    \item \textbf{Missed Insights in Bias Intensity:} The system currently outputs a binary bias intensity score (\texttt{High} or \texttt{Low}), which may oversimplify the complexity of opinionated statements. This approach could overlook the nuances of bias severity, limiting the depth of insights provided by the system and its capacity for complex justifications.
\end{itemize}

\subsection{Future Work}
Our research presents several paths where future work can be introduced to improve the effectiveness and applicability of our bias detection framework:
\begin{itemize}
    \item \textbf{Specific bias evaluation:} To introduce a more comprehensive detection system, future work could focus on producing a more detailed assessment of biased statements and providing better explainability. For example, a bias determiner agent could be implemented in the framework to classify for specific social or cultural biases rather than solely distinguishing between biased and unbiased text.
    \item \textbf{Percentage scoring for bias:} Transitioning from binary bias scores to a continuous or percentage-based scale would enable a more detailed assessment of bias intensity. This change could provide users with richer interpretability and a clearer understanding of the system’s decision-making process.  
    \item \textbf{Contextual bias detection:} Incorporating context-aware models or Retrieval-Augmented Generation (RAG) systems could improve the framework’s ability to detect context-specific biases \cite{lewis2021}. These enhancements would allow for more detailed justifications grounded in relevant contextual information.  
    \item \textbf{Broader dataset application:} To improve generalizability, the framework should be evaluated on datasets from diverse domains such as medicine, law, and journalism, where objectivity and fairness are critical. This would help assess the system’s adaptability to varying linguistic and contextual nuances.  
\end{itemize}

\section{Ethical Considerations}
The development of our bias-detection framework requires careful attention to transparency, fairness, and the limitations inherent in bias evaluation. As biases are often subjective and embedded in the training data, our framework’s accuracy depends on the diversity and representativeness of the data sources used. Ensuring inclusivity in training data and incorporating broader contextual information are necessary steps to address this challenge. Additionally, while the framework provides justifications for its classifications, these explanations may not always fully capture the nuances of complex biases, highlighting the need for more detailed and comprehensive rationalizations. By addressing these concerns, we aim to create a more equitable and transparent system that builds trust and supports ethical AI development.

\section{Conclusion}

This paper introduced a multi-agent framework for bias detection that combines fact–opinion classification, bias verification, and concise justification. Across 1,500 statements from the WikiNPOV dataset, the framework significantly outperformed a zero-shot baseline ($p < 10^{-6}$), demonstrating the impact of a structured, agent-based pipeline for improving both accuracy and interpretability. Additionally, experiments with multiple LLMs, including GPT-4o and Claude 2.1, confirmed the system’s adaptability. By generating transparent justifications for each classification, our method offers practical advantages in domains requiring trustworthy AI-driven decisions. Future directions include expanding the pipeline to multilingual contexts, integrating external knowledge bases, and adding confidence calibration to further bolster reliability and user trust. Ultimately, this framework contributes to AI-driven systems aiming to promote fairness and accountability, where structured, transparent bias detection can function as a protection in the broader pursuit of trustworthy AI.

\bibliography{aaai25}

\end{document}